# Retrieval Augmented Generation Framework for the Nepali Legal Domain Question Answering

Samir Wagle[1], Abiral Adhikari[1], Reewaj Khanal[1], Batsal Bhandari[2], Prashant Manandhar[1], Praveen Acharya[3], and Bal Krishna Bal[1]

[1]*Information and Language Processing Research Lab*
*Kathmandu University , Dhulikhel, Kavre, Nepal*
ORCIDs: 0009-0002-5126-1107 (SW), 0009-0006-8198-2975 (AA), 0009-0007-4005-271X (RK)
Emails: prashantmanandhar2002@gmail.com, bal@ku.edu.np

[2]*School of Law*
*Kathmandu University*, Dhulikhel, Kavre, Nepal
Email: bhandaribatsal@gmail.com

[3]*Dublin City University*
Dublin, Ireland
Email: acharyaprvn@gmail.com

***Abstract*—Legal domains in high-resource languages like English have widely adopted artificial intelligence for legal question answering. However, data scarcity in low resource languages such as Nepali has limited the training of large language models on Nepali legal texts. This study presents the first application of a Retrieval Augmented Generation based model for Nepali legal question answering using case laws extracted from the Nepal Kanun Patrika digital archive. Using BM25 on chunked documents, the approach achieved a top precision at one of 91 percent, and up to 75 percent with the multilingual E5 large model. Evaluation of generated answers showed 74 percent groundedness, 85 percent truthfulness according to an automated judge model, and 84 percent human evaluated truthfulness when using BM25 document retrieval, with a 92 percent successful answer generation rate. These results demonstrate that the RAG pipeline can effectively address the gap in legal question answering for low resource languages and provide a foundation for reliable AI systems in the Nepali legal domain.**

## I. Introduction

The Nepali legal system, entrenched in a hybrid of common and civil law traditions and functioning primarily in the Nepali language, faces significant challenges in information accessibility, consistency, and linguistic inclusivity. Nepali legal texts are often fragmented, poorly digitized, and written in archaic terminology, making manual research time-consuming and inefficient. The dual challenge of the scarce availability of centralized legal documents, the abstruse and jargon-laden nature of the documents hinders public engagement in the legal domain.

The imperative to enrich public participation within the legal domain has given rise to the critical need for AI systems with question-answering capabilities that can engage the general public and legal professionals alike. This endeavor is further complicated by Nepali being a low-resource language, which has precluded the creation of a specialized, fine-tuned legal LLM currently infeasible.

This data-scarce environment necessitates a strategic departure from conventional fine-tuning methodologies, making the Retrieval-Augmented Generation (RAG) model a particularly suitable architecture. The RAG model circumvents the need for extensive, manually annotated training datasets. The dynamic retrieval of relevant documents at the point of inference to enrich the generative process ensures the optimal use of limited resources while anchoring its output in verifiable legal text.

To the best of our knowledge, this research presents the first systematic application of RAG based question answering pipeline unique to Nepalese case laws, specifically landmark Supreme Court rulings selected from NKP (Nepal Kanun Patrika) [1]. We examine two distinct methodologies - sparse retrieval using BM25 on document and chunk level (splitting documents into fixed-length passages) and dense retrieval using three different embedding models - multilingual-e5-large, multilingual-e5-base, and LaBSE.

The contributions of this study include:

- Compiled and curated the first public Nepali Supreme Court case law dataset from NKP archive, for RAG

*(Samir Wagle, Abiral Adhikari, Reewaj Khanal, Batsal Bhandari, Prashant Manandhar, Praveen Acharya, and Bal Krishna Bal contributed equally to this work.) (Corresponding author: Abiral Adhikari.)*

[1] https://nkp.gov.np/

based QA application.

- Established the first baseline application of a standard RAG pipeline for Nepali legal QA.
- Conducted the first comparison of sparse (BM25) versus dense retrieval methods for this domain.

## II. Related Works

The field of Question Answering (QA) has undergone a major transformation as a result of Large Language Models (LLMs), which have illustrated strong performance in understanding and generating human-like text. While LLMs are powerful tools for extraction and generation, applying them in technical high-stakes domains such as law requires a more sophisticated strategy to ensure correctness and to reduce the risk of factual hallucination. A more recent survey by [1] provides a broad discussion of legal QA systems, giving a comprehensive and thorough overview of the methods, datasets, and evaluation approaches that comprise the current state of the art. Their survey also highlighted the unusual nature of legal QA systems, similar to [2], whose research cover the scope of legal Natural Language Processing (NLP) as a whole, including the major tasks, state of models that have been developed to deal with these tasks, and major challenges still faced in this research area, such as the lack of data and the insistence of legal jargon.

Effective legal QA systems rely on robust Information Retrieval (IR). The foundations of legal IR were laid out by [20], which discussed the issue of ambiguity and term variations in legal texts. The systems developed in the early days of legal QA typically used classical IR methods and principles based on [20]'s findings. To illustrate, [6] employed domain-based indexing and query expansion for a legal search system and demonstrated the classical relevance of conventional IR models.

More recently, efforts have focused on legal case retrieval, a key sub-task for most legal QA systems. The comprehensive survey conducted by [10] summarized the development of legal case retrieval from the prior keyword search systems to the recent semantic models while also providing an overview of prominent approaches to legal case retrieval, major benchmarks, and difficult challenges that remain problematic for all systems, including the role of drawing conclusions given the different underlying legal systems in each jurisdiction.

To address the shortcomings of standard LLMs, especially their tendency to create nonsensical information, RAG has emerged as a robust approach [14]. RAG provides stronger reliability than LLMs by using a set of retrieved documents as the grounding for its responses to provide contextual relevance and maintain factual accuracy. Allowing for a retriever and generator illustrates a straightforward method of creating accountable and explainable QA systems [12].

The grounded generation from RAG has consequences within the legal domain, with misinformation having potentially very negative consequences. By linking generated answers to real statutes and case laws, RAG systems can generate legally coherent and verifiable outputs. There is a lot of innovation going on in this space. The law-guided generative approach put forth by [19], fuses judgment prediction and retrieval of documents to augment both the effectiveness of retrieval and legal reasoning. Similarly, [23] provides evidence for the power of hybrid architectures by combining transformer-based LLMs with Graph Neural Networks to model the legal citation relationships and achieve state-of-the-art results on the COLIEE 2025[2] benchmark.

While methods of this level of sophistication move the goal posts in what can be achieved, the component pertaining to retrieval is still important. Legal document identification, where the sparseness and frequency of domain jargon and phrases make term-frequency-based sparse retrieval still appropriate [21]. The performance of BM25 on recall-based benchmarks like the TREC Legal Track[3] and COLIEE speaks to its ability across the legal domain [25].

While the international interest in legal NLP continues to grow, studies have generally focused on formal, high-resource languages, and surprisingly few studies have explored low-resource languages, especially lower-endowed Asian languages like Nepali. Many recent advances in RAG-based legal Q&A systems have emerged for other morphologically rich, under-resourced Asian languages such as Vietnamese [3] and Thai [18], with evidence showing robust findings.

Although the Nepali legal domain is uniquely under-researched and used, this is primarily due to a lack of representative annotated legal corpora, little digitalization of case law, and no standardized NLP benchmarks. Essentially, in this study, we are working in a data-scarce environment, and fine-tuning LLMs is impractical, whereas context-bounded generation of RAG has made use of otherwise impossible NLP processes practical and powerful. Also, we are not alone in data-scarcity issues: as an example, [15] has recently used synthetic data to fine-tune legal LLMs, but RAG is easier as there is little to no pre-training needed.

There has not been any previous effort, to the best of our knowledge, that has systematically utilized or evaluated retrieval strategies, whether sparse, dense, or a hybrid of both, in an RAG pipeline for the Nepali legal domain, meaning that the opportunity for baseline foundational research is both urgent and needed. Therefore, we propose a first-of-its-kind RAG-based framework for Nepali legal question answering, and our work will compare a classical sparse retriever with a modern dense retriever on a curated corpus of Nepali case law.

[2] https://coliee.org/overview

[3] https://trec.nist.gov/

Table I
SAMPLE OF DOCUMENT COLLECTION FOR INDEXING

| Field | Value |
|---|---|
| Document ID | 3cc4af57-ab5d-415c-9374-947dee546722 |
| Chunk ID | D7000_C12 |
| Document Index | 7000 |
| Chunk Number | 12 |
| Subject | कर्तब्य ज्यान । |
| Case Number | निर्णय नं. ८३१५ |
| Decision Date | फैसला मिति : २०६६/१०/०४ |
| Source | nkp.gov.np/full_detail/3303 |
| Page Content[a] | दन अदालत पाटनको मिति २०६०।८।१ गतेको फैसला ... विपक्षी झिका |

[a]Truncated for brevity. Full content stored in vector index.

## III. DATASET AND METHODOLOGY

### A. Dataset

A total of 10,320 legal documents corresponding to individual cases were extracted from the NKP (Supreme Court) digital archive through web scraping, implemented using a Python script in combination with the BeautifulSoup library for HTML content parsing. The collected documents' decision dates ranging from 1992 AD to 2022 AD. The case documents include selected monumental judicial decisions by the Supreme Court of Nepal comprising 2847 distinct case types, including homicide, rape, private civil case, succession, writs such as habeas corpus, mandamus, certiorari, prohibition, etc. The collected data was preprocessed with line breaks, excessive whitespace, empty document removal, and Unicode Normalization Form Canonical Composition (NFC) normalization for consistency. Even with this preprocessing, a few anomalies, such as random unwanted characters due to inconsistent formatting during the historical data digitization process, were observed.

The cleaned documents were then segmented into chunks with a maximum of 500 tokens with 10% overlap between consecutive chunks to maintain contextual continuity across segments. The final dataset curated for retrieval contained 10,265 unique documents, which were divided into 110,626 chunks. The divided chunks looks like as shown in Table I

To evaluate the model's retrieval performance, a test set comprising 100 queries were deliberately curated by legal domain scholars under the supervision of Professors from the Law Faculty of Kathmandu University School of Law. This design approach of prioritizing legal correctness and annotation fidelity over scale ensures each query strictly corresponds to a single relevant case document. The contemplation behind fact-dependent questions based test set preparation, was to ensure verifiability from legal context and to prevent fallout due to hallucination.

The fact-based objective queries comprised 79% of the set, focusing on specific facts or definitions or legal provisions (87% of the objective category) and identification of involved parties (13%). The remaining 21% of queries were subjective, assessing the model's capacity to retrieve information on legal procedure and reasoning (58% of the subjective category) and the rationale behind judicial decisions (42%).

### B. Retrieval Method

*1) Sparse Retrieval:* BM25 was the selected method for the sparse retrieval due to its strong legal domain performance in the study by [13]. In this study, we implemented BM25 twice for performance comparison, as follows:

- **BM25_Docs:** BM25 retrieval was implemented on the complete document.
- **BM25_Chunks:** BM25 was implemented after dividing documents into chunks with a maximum of 500 tokens, including a 10% overlap between chunks.

All the documents and chunks were indexed and retrieved using the BM25Retriever provided by Langchain[4].

*2) Dense Retrieval:* In addition to the sparse retrieval, the dense retrieval method was evaluated with three leading models on the chunked documents:

- **Multilingual-e5-large**, with a maximum input token size of 512 and a dimension size of 1024, was selected for its state-of-the-art zero-shot performance on MTEB, BEIR, and TAT-QA, the highest among multilingual dense retrievers [25].
- **Multilingual-e5-base:** with a maximum input token size of 512, was chosen for its efficient 768-dimension size representation and strong zero-shot retrieval performance in multilingual tasks [25].
- **LaBSE** [9] with a maximum input token size of 512, generating embeddings of dimension size 768, was selected due to its strong performance in multilingual retrieval and question answering across over 100 languages.

Based on the common input size for all the models, the entire document corpus, split into chunks with a maximum of 500 tokens, was encoded and indexed using FAISS for fast retrieval.The 500-token chunk size was specifically chosen to remain within the 512-token limit of the embedding models, with the 10% overlap ensuring that no critical legal context was lost at the chunk boundaries. At query time, queries were similarly encoded, and cosine similarity was used to identify and rank the top-K candidate chunks from the FAISS index. The top-K chunks were retrieved as the candidates for the context document selection.

[4]https://www.langchain.com/

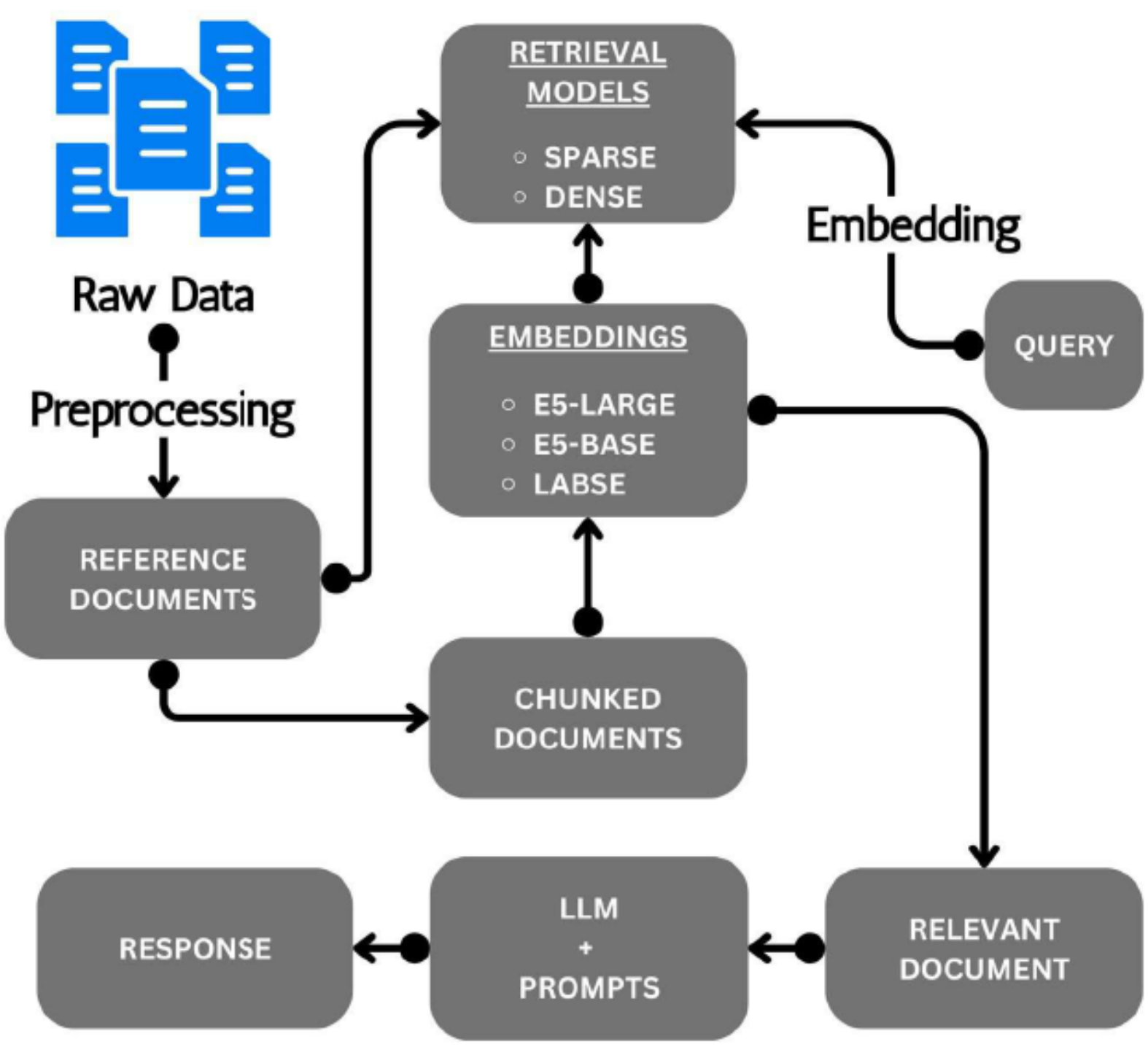

Figure 1. Retrieval-Augmented Generation (RAG) pipeline of the experiment.

Table II
SAMPLE QUESTION AND ANSWER GENERATED FROM BM25 DOCS, ALONG WITH GROUND TRUTH.

| Question | Ground Truth | Generated Answer |
|---|---|---|
| सर्वोच्च अदालतले, सम्झौता हुँदाको अवस्थामा धन बहादुर गुरुङ्ग प्रिन्सिपल रहेको र स्कुलको छाप लागेको देखिएकोले कसको तर्फबाट करार भएको देखिने ठहर गर्यो? | सर्वोच्च अदालतले धन बहादुर गुरुङ्ग प्रिन्सिपल रहेको र स्कुलको छाप लागेको देखिएकाले विद्यालयको तर्फबाट करार भएको ठहर गर्यो। | सर्वोच्च अदालतले धन बहादुर गुरुङ्ग प्रिन्सिपल रहेको र स्कुलको छाप लागेको देखिएकोले स्कुलको तर्फबाट करार भएको देखिने ठहर गर्यो। |

### C. Retrieval Evaluation

The retrieval performance was assessed using three standard metrics tailored for the test set-Precision@1, Hit@K, and Mean Reciprocal Rank (MRR@k) to collectively measure both the presence and the ranking quality of retrieved results, as suggested in [11].

For our test set, where the query has a single ground truth document, Precision@k would never exceed the 1/k value [22], rendering it useless to be incorporated for our evaluation. In the same context, Recall@k inherently becomes Hit@k, leading to its exclusion. The redundant evaluation of Hit@1 and MRR@1 was also avoided with the calculation of Precision@1, which in the given testing context conveys the same significance. Each retrieval method was evaluated with these metrics across multiple cut-off thresholds (k = 1, 3,5,10,15,20) before proceeding to answer generation, along with their latency (measured in seconds) to evaluate the tradeoff between performance and computation efficiency.

### D. Generation Method

The contexts for the generation component were retrieved by three different methods. The first-ranked document was selected from BM25_Docs and e5-large retrieval, which had the best performance and latency results. In the Count Method, the most frequent document was chosen after aggregating the top 5 documents retrieved by BM25_Docs, BM25 _Chunks, and e5-large.This ensemble approach was designed to test if consensus across sparse and dense retrievers could improve grounding by selecting the most 'agreed-upon' document, though results later showed it underperformed individual methods. Since legal texts often depend on the full document for proper understanding, we selected the entire document corresponding to the retrieved chunk, even though the retrieval was done at the chunk level, to ensure accurate answers.

To improve contextual accuracy, we applied a refusal strategy in all methods to avoid hallucination. This instructs the language model to respond only when sufficient legal grounding is present in the retrieved context. For the generation part, we applied a two-step prompting approach. In the first step of evidence extraction, the model identifies verbatim legal sentences from the context that support the answer. In the second step, it generates the final answer based on that evidence. If no relevant content is found, the model is instructed via a prompt to refuse answering with a standard fallback response. This strategy helps preserve factual integrity by preventing the model from fabricating answers when evidence is missing.

For the final answer generation part based on the earlier retrieved case details, we chose OpenAI's o3 model. This model was chosen for its strong retrieval performance and reasoning ability across multilingual QA benchmarks, even without domain-specific adaptation, as shown by [4]. In our experiment in generation, we deliberately maintained a moderate temperature (0.5) and full top-p (1.0) to balance factuality and fluency while capping output length to 2048 tokens. However, we observed that the prompt design (two-step strategy) had a more profound impact than generation parameters alone. These decisions demonstrate that the success of the generation pipeline depends on retrieval quality.

### E. Generation Evaluation

The generated answer's quality was evaluated using three key metrics: Groundedness, Truthfulness, and Human Evaluation. The factual alignment based on retrieved legal sources was confirmed through groundedness and truthfulness evaluation. Following best practices from [7], LLMs can perform factual consistency checks by decomposing outputs into discrete claims and verifying each claim against grounding documents, when prompted appropriately.

Human Evaluation was used to assess the clarity, accuracy, and usefulness of the model's output. This follows the guidance of [8], who emphasized the essential role of human judgment in evaluating LLM-generated responses, especially in domain-sensitive and high-stakes contexts like law.

To perform this evaluation, we designed prompts for Groundedness and Truthfulness for the LLM-as-a-judge approach using Gemini 2.5 Pro. In the groundedness, the model was instructed to examine the generated answer, identify the factual claims and verify that each claim was explicitly supported by the retrieved legal document. The prompt explicitly discourage inference, external knowledge, or biased reasoning beyond the provided context. A generated answer was considered grounded only when all of its claims could be directly verified from the source document. Else, it was identified as non-grounded.

For the evaluation of truthfulness, the model compared generated answer with an annotator curated reference answer and examined whether both answers expressed the same core legal facts, allowing for minor word or phrase differences. The prompt was written in such a way that if there were any contradictions, missing key details, or factual errors, the answer should be marked as non-truthful. In both evaluations, the model was limited to simple yes-or-no (binary) responses. This approach ensured that the results were consistent, easy to interpret, and scalable especially for legal question-answering tasks where clarity and precision are essential.

## IV. Results

### A. Retrieval Performance

The performance and computational efficiency of five retrieval strategies evaluated using Precision@1, Hit@k, and MRR@k for different values of k are presented in Table III and Table IV, respectively. Though the experiment was conducted with values of k=1,3,5,10,15,20 , the results of k=10,15,20 are not mentioned in Table III as there was no significant improvement in performance.

Sparse retrieval, particularly BM25_Chunks, demonstrated decisively superior performance, achieving a Precision@1 of 0.91. This significantly outperformed the best dense model, multilingual-e5-large (Precision@1 = 0.75), while other dense models showed considerably weaker results. However, this higher performance came with a significant trade-off in latency, which scales up with a larger number of chunks. The top-performing BM25_Chunks was the slowest method by a large margin, at 1.2 seconds per query. In contrast, the fastest models were the lowest-scoring LaBSE and multi-lingual-e5-base ( 0.04 s/query).

In the case of sparse retrieval methods, BM25_Chunks had only slightly better performance against very high retrieval time compared to BM25_Docs. The MRR@k metrics of the models also show that the models with better performance have the correct document retrieval within the top 3 documents. So, the retrieval methods that had balanced performance and computational efficiency: BM25_Docs (sparse retrieval) and multilingual e5-large(dense retrieval), were selected for context retrieval for generation purposes.

### B. Generation Performance

These methods were applied to a held-out test set of 100 legal questions. Table V summarizes how many answers

Table III
Retrieval Performance Comparison for $k = 1, 3, 5$

| Method | P@1 | Hit@3 | Hit@5 | MRR@3 | MRR@5 |
|---|---|---|---|---|---|
| e5_large | 0.75 | 0.78 | 0.80 | 0.762 | 0.766 |
| BM25_Docs | 0.88 | 0.95 | 0.96 | 0.913 | 0.916 |
| BM25_Chunks | 0.91 | 0.94 | 0.95 | 0.920 | 0.923 |
| e5_base | 0.38 | 0.45 | 0.50 | 0.410 | 0.422 |
| LaBSE | 0.22 | 0.28 | 0.31 | 0.245 | 0.252 |

Table IV
Average Query Time per Retrieval Method

| Method | Time per Query (seconds) |
|---|---|
| e5_large | 0.0681 |
| BM25_Docs | 0.1042 |
| BM25_Chunks | 1.2003 |
| e5_base | 0.0436 |
| LaBSE | 0.0434 |

were successfully generated based on the quality of the retrieved context.

Among the methods that were evaluated, BM25_Docs achieved the best performance with an impressive 92% accuracy in answer generation. This highlights its strong retrieval capability, which supports accurate responses within legal contexts. In contrast, the Count method's performance was poor, generating only 78 answers. Its dependence on basic term frequency likely led to the omission of important but less frequent legal information, resulting in a higher number of unanswered queries.

In terms of answer quality, BM25_Docs also achieved the highest groundedness score of 74%. This higher value indicates that there is a strong alignment between the retrieved content and generated answers. Furthermore 85% truthfulness score indicates its consistent factual accuracy. In contrast, the Count method fell short on both measures, with 61% groundedness score and 72% truthfulness score highlighting the limitations. The human truthfulness evaluation supports the results generated with the "LLM-as-a-Judge" approach, resulting in a higher error rate in truthfulness for count and e5-large method when compared to BM25_Docs. These differences make it clear that having a strong retrieval system like BM25 is crucial when it comes to generating reliable, relevant answers in legal question answering tasks.

## V. Discussion

The results of the study indicate that sparse retrieval using BM25 outperforms dense retrieval with out-of-shelf embedding models such as multilingual-e5-large for factual queries with formulaic legal language, promoting greater lexical match. This finding is in agreement with prior research by [16].

The results of the experiment offer the insight that the model demonstrates only marginal improvements in retrieval performance with increasing values of k, indicating information saturation due to the nature of the queries. The retrieval effectiveness of the study's retrieval model reached Precision@1 of 75% for e5-large, 88% for BM25_Docs, and 91% for BM25_Chunks, which are modest results for the curated test set.

The lower performance of dense retrieval here is consistent with findings from prior studies such as [5]. This weakness is partially attributed to the lack of domain-adaptive pretraining of the dense retrieval models on Nepali legal text. Furthermore, the lexically constrained and highly formulaic nature of Nepali legal documents exhibiting minimal variations in statuary expression, amplifies the effectiveness of sparse lexical matching making exact-word overlap more informative than semantic approximation.

The acknowledgement of the trade-off between performance and efficiency by [18] motivated us to discard even the best-performing BM25_Chunks method. Although BM25_Chunks provided a slight performance improvement of 3.4% at the cost of 11 times higher latency during retrieval compared to BM25_Docs method. The higher latency of BM25_Chunks is solely due to the larger number of chunks: 110,626 compared to the mere 10,265 number of documents, which is 10.78 times less. Given this disproportionate retrieval overhead over marginal performance gain, BM25_Docs presents a more efficient and viable retrieval strategy.

The performance difference noted between dense retrieval models indicates superior performance of e5-large over e5-base and LaBSE. This superiority of e5-large can largely be attributed to its deeper 24-layer architecture, higher 1024 embedding dimension size, and larger model capacity of 570 M parameters when compared to e5-base and LaBSE with 12-layer architecture, lower 768 embedding dimension size, and lower model capacity with 278 and 471 M parameters, respectively. This leads the e5-large to learn more nuanced semantic representation [9], [25]. The inherent fragmentation of legal information across multiple sections, clauses and references within a single document renders selective retrieval from smaller semantic chunks ineffective in the legal domain , despite being an effective strategy in general-domain retrieval tasks. It is to be noted that with high chances of performance improvement, experiments with hybrid retreival and re-ranking strategies are imperative in the future.

In the answer generation part, the OpenAI GPT o3 model was implemented and evaluated based on answer accuracy, groundedness, and truthfulness. The generation results strongly correlated with retrieval quality: BM25-based retrieval achieved the highest answer generation capability (92%), groundedness (74%), and truthfulness (85%). The reason behind better performance is because of the two-step prompt strategy that guides the model to first extract legal evidence and then generate answers strictly based on that evidence. This approach effectively minimizes hallucinations, as confirmed by the empirical absence of fabricated answers in our outputs, and also mentioned in a previous study by [26], [27].

Table V
Generation Performance Comparison

| **Method** | ***Truthfulness (%)*** | ***Groundedness (%)*** | ***Successful Answers (out of 100)*** |
|---|---|---|---|
| BM25_Docs | 85 | 74 | 92 |
| e5_large | 78 | 62 | 81 |
| count | 72 | 61 | 78 |

Furthermore, this strategy instructs the model to explicitly decline answering when insufficient evidence is available, preserving factual integrity and preventing misinformation, which is a critical factor in legal domains. The Nepali-language system prompt also enhances domain relevance and contextual understanding, allowing the model to produce coherent and culturally appropriate answers. The choice of GPT-o3, despite lacking domain-specific fine-tuning, demonstrates that retrieval quality influences the generation outcomes [17].

In this study, lexically precise and comprehensive BM25 retrieval contexts provided more truthful and grounded answers compared to dense or count-based retrieval methods. In addition to automated evaluation, human evaluation was incorporated in our experiment, which is essential for evaluating the LLM's generation performance. However, with human evaluation, the introduction of human biases is unavoidable, which can lead to different outcomes compared to automated methods like LLM-as-a-Judge, as observed in the generation evaluation in our study. As noted by [24], the mitigation of such biases that arise from subjective interpretation, task ambiguity, and inconsistency across annotators requires the use of well-defined criteria such as clarity, accuracy, and usefulness when setting up evaluation outputs in sensitive domains like legal QA.

## VI. Conclusion

This study introduced the first Retrieval-Augmented Generation (RAG) framework tailored for Nepali legal question answering using Supreme Court case law from the NKP digital archive. In a low-resource setting with limited annotated data, our pipeline demonstrates that retrieval quality directly impacts generation quality, with sparse retrieval, particularly BM25-based methods, proving far more effective than dense retrieval using off-the-shelf multilingual embeddings.

Among the retrieval methods evaluated, BM25 _Chunks achieved the highest retrieval accuracy (Precision@1 = 91%), outperforming the best-performing dense method, multilingual-e5-large (75%). However, BM25_Chunks was avoided in generation due to its higher latency (1.2s/query). Instead, BM25_Docs was selected due to the latency tradeoff, offering nearly comparable performance (Precision@1 = 88%) with lower latency. The best-performing sparse retrieval using BM25_Chunks with 91% accuracy shows that, for our experiment setup, sparse retrieval is more effective in the Nepali legal textual setting.

In the generation phase, the answers were generated using top-performing retrieval methods. Both LLM-as-a-Judge and human evaluation were used to evaluate factual integrity and clarity of generated answers. Here, BM25_Docs led to the best overall results with 92% answer generation rate, 85% truthfulness, and 74% groundedness score using the LLM-as-a-Judge approach, and 84% on human truthfulness evaluation.

The LLM-as-a-Judge approach showed similar results to human evaluation, highlighting the effectiveness of automated evaluation. However, some differences were seen mainly due to human biases and subjectivity being a known challenge in legal QA tasks. This experiment, though carried out meticulously, was observed to have few limitations. The reliance on a test set of 100 expert-annotated queries, is one such limitation which hinders broad statistical generalization.

However, acknowledging the fact that construction of legally valid and fact-dependent Nepali legal questions requires substantial expert involvement along with limitation of scarce test set, we plan to actively expand the benchmark with additional expert-curated queries. Following which our future releases will incorporate a larger and more diverse test- set.

Additionally, the state-of-the-art hybrid retrieval strategies and re-ranking algorithms are to be explored in the future for performance enhancement. Moreover, the data was limited to case laws from the NKP digital archive, which may not represent the full Nepali Legal domain. The solutions to the mentioned limitations will be included in future works to improve the applicability and performance of the proposed pipeline. These future works will include expansion of the dataset, exploration of hybrid and reranking strategies, and establishment of a proper Nepali legal benchmark for baseline evaluation.

## Appendix

In this appendix, we provide the exact prompts used in our experimental pipeline for answer generation, truthfulness evaluation, and groundedness evaluation.

### A. Prompt 1: Answer Generation

This prompt generates a precise, formal, and factually accurate answer to a legal question in Nepali, strictly based on the provided context.

**Prompt Text:**

*Role: You are an expert Nepali legal assistant that answers strictly using provided legal context. Never speculate beyond the given information.*

*Rules: 1. Answer ONLY using verbatim phrases from the provided context in Nepali. 2. If the question cannot be answered from context, respond:* ”यस प्रश्नको उत्तर दिन उपलब्ध प्रसंगमा आधारभूत जानकारी छैन।” *3. Preserve original legal terminology (e.g.* ”धारा १८८”, ”अभियोग दायर”*).*

*Input Format: { ”context”: ”[Retrieved legal chunks from CSV]”, ”question”: ”[User’s Nepali legal query]” }*

*Output Requirements: 1. Language: Strictly Nepali (with Devanagari script). 2. Length: 1-3 sentences maximum. 3. Format: Plain text (no JSON/Markdown). 4. Must include at least one exact phrase from context.*

*Quality Control: Answers will be programmatically verified for substring matches with context, absence of external knowledge, and compliance with Nepali legal terminology.*

*Example: Input Context:* ”देवानी संहिता २०७४ को धारा २७३ : जमिनको हकमा प्रतिकूल भोगाधिकार कायम हुन ३० वर्ष आफ्नै सरह भोग गरेको हुनपर्छ।” *Question:* ”जमिनको हकमा प्रतिकूल भोगाधिकार कायम हुन कति वर्ष कट्नु आवश्यक हुन्छ?” *Correct Output:* ”जमिनको हकमा प्रतिकूल भोगाधिकार कायम हुन ३० वर्ष कट्नु आवश्यक हुन्छ।”

—

*context: {context}*
*question: {question}*

### B. Prompt 2: Truthfulness Evaluation

This prompt determines whether a generated answer is factually consistent with a ground-truth answer.

**Prompt Text:**

*You are an impartial evaluator. Your task is to verify whether the generated answer is factually consistent with the provided ground truth answer.*

*Instructions: If the generated answer and the ground truth express the same meaning, even if phrased differently, consider them truthful / equivalent. If the generated answer contains factual inaccuracies or contradicts the ground truth, consider it not truthful. If the generated answer exactly contains any of these cases:* ”यस प्रश्नको उत्तर दिन उपलब्ध प्रसंगमा आधारभूत जानकारी छैन।” *then assign Truthfulness Score = 0 regardless of other factors.*

*Focus on semantic equivalence and factual correctness, not identical wording. Answer should be 0 or 1.* —

*The ground truth answer is {ground_truth_answer}*

—

*Generated Answer:*

—

*{generated_answer}*

—

### C. Prompt 3: Groundedness Evaluation

This prompt verifies that all claims in the generated answer are fully supported by the provided context.

**Prompt Text:**

*Please act as a meticulous fact-checker. Your task is to determine if the ’Answer’ below is fully supported by the provided ’Context’.*

*Read the ’Context’ and then the ’Answer’ carefully.*

*- If all factual claims in the ’Answer’ can be directly and explicitly verified from the ’Context’, respond with only the number ’1’.*

*- If any part of the ’Answer’ contains information that is not present in the ’Context’ (even if the information is plausible or true in the real world), respond with only the number ’0’.*

*Context:*

—

*{context.strip()}*

—

*Answer:*

—

*{answer.strip()}*

—

## Data and Code Availability

To support the reproducibility of this research, the complete experimental codebase, scraping scripts, and the curated dataset of Supreme Court case laws are publicly accessible. In accordance with the double-blind review process, these resources have been anonymized and can be accessed at: ”https://github.com/SamirWagle/Retrieval-Augmented-Generation-Framework-for-the-Nepali-Legal-Domain-Question-Answering” . Upon acceptance, the resources will be deanonymized and hosted on a permanent public repository.